\documentclass[conference]{IEEEtran}
\IEEEoverridecommandlockouts
\usepackage{cite}
\usepackage{amsmath,amssymb,amsfonts}
\usepackage{algorithmic}
\usepackage{graphicx}
\usepackage{textcomp}

\usepackage{color,soul}

\usepackage[table]{xcolor}
\usepackage{tikz}
\usetikzlibrary{patterns}

\usepackage{multirow}
\usepackage{caption}
\usepackage{subcaption}

\usepackage{hhline}
\usepackage{bm}
\usepackage{tablefootnote}
\usepackage{xfrac}
\usepackage[pagebackref=false,breaklinks=true,colorlinks,bookmarks=false]{hyperref}

\usepackage{setspace}

\usepackage{fancyhdr}
\fancypagestyle{firststyle}
{
   \fancyhf{}
   \fancyhead[C]{
    \textcolor{red}{
       The content of this paper was published in EUSIPCO 2022. If you find the paper useful, please cite the following paper: \\ Ž. Babnik and V. Štruc, "Assessing Bias in Face Image Quality Assessment," 2022 30th European Signal Processing Conference (EUSIPCO), IEEE 2022, pp. 1037-1041.}
    }
}

\def\BibTeX{{\rm B\kern-.05em{\sc i\kern-.025em b}\kern-.08em
    T\kern-.1667em\lower.7ex\hbox{E}\kern-.125emX}}
\renewcommand{\baselinestretch}{0.97}
\begin{document}

\title{Assessing Bias in Face Image Quality Assessment
\thanks{Supported by the ARRS Research Program P2--0250 (B) as well as the ARRS Junior Researcher Program.}
}

\author{\IEEEauthorblockN{Žiga Babnik}
\IEEEauthorblockA{\textit{University of Ljubljana, Slovenia} \\
ziga.babnik@fe.uni-lj.si\vspace{-6mm}}
\and
\IEEEauthorblockN{Vitomir Štruc}
\IEEEauthorblockA{\textit{University of Ljubljana, Slovenia} \\
vitomir.struc@fe.uni-lj.si\vspace{-6mm}}
}


\maketitle
\thispagestyle{firststyle}

\setstretch{0.93}

\begin{abstract}
Face image quality assessment (FIQA) attempts to improve face recognition (FR) performance  by providing additional information about sample quality. 
Because FIQA methods attempt to estimate the utility of a sample for face recognition, it is reasonable to assume that these methods are heavily influenced by the underlying face recognition system. Although modern face recognition systems are known to perform well, several studies have found that such systems often exhibit problems with demographic bias. It is therefore likely that such problems are also present with FIQA techniques. To investigate the demographic biases associated with FIQA approaches, this paper presents a comprehensive study involving a variety of quality assessment methods (general-purpose image quality assessment, supervised face quality assessment, and unsupervised face quality assessment methods) and three diverse state-of-the-art FR models. 
Our analysis on the Balanced Faces in the Wild (BFW) dataset shows that all techniques considered are affected more by variations in race than sex. While the general-purpose image quality assessment methods appear to be less biased with respect to the two demographic factors considered, the supervised and unsupervised face image quality assessment methods both show strong bias with a tendency to favor white individuals (of either sex). In addition, we found that methods that are less racially biased perform worse overall. This suggests that the observed bias in FIQA methods is to a significant extent related to the underlying face recognition system.

\end{abstract}

\begin{IEEEkeywords}
bias, bias estimation, demographics, biometrics, face recognition systems, face image quality assessment 
\end{IEEEkeywords}

\section{Introduction}\label{sec:introduction}


Modern Face Recognition~(FR) systems are capable of achieving excellent results on large datasets containing images of varying characteristics, such as pose, illumination or occlusions. Yet many challenges still exist that prevent such performance to carry over to real-world scenarios \cite{grm2018strengths}. 


Face Image Quality Assessment~(FIQA) aims to assist FR models in achieving better performance by providing additional biometric sample quality information. Unlike standard Image Quality Assessment~(IQA), which is tightly connected to human (visual) quality perception, 
FIQA techniques focus on estimating the utility of the input samples for FR tasks. As such, the quality scores obtained from FIQA methods may not  directly reflect visual quality but account for different image characteristics that may have an impact on recognition performance. One of the key problems of FR models not explicitly addressed by FIQA techniques is bias. FR models have been shown to exhibit different performances for different demographic groups~\cite{frbias1, frbias2}. These performance differentials (or bias) have an impact on the \textit{fairness} of FR models and have recently been at the core of many research efforts. As FIQA techniques are intended to capture the utility of face images for face recognition, 
it seems likely that bias-related issues are also present with this group of techniques. Understanding the behavior of FIQA methods especially in conjunction with selected FR models is, therefore, critical for the trustworthiness of face recognition technology and its perceived fairness in the general public. 

While many studies have been conduced on the topic of bias in FR tasks~\cite{frbias1, frbias2, frbias3, frbias4}, exploring differences between race, sex and age groups, not much research has been done on the topic of bias in FIQA methods, with one notable exception. In \cite{fiqabias}, Terhörst \textit{et al.} investigated bias in FIQA techniques by looking at how gradually increasing the number of rejected images, due to a poor quality score, changes the proportions across different demographic groups. The main assumption of this work was that fair and unbiased techniques should exclude samples from different demographic groups in an equal manner. In this paper, we built on the work in~\cite{fiqabias} but take a step further and explore FIQA methods and FR models within a joint framework. Thus, we are not interested in the (demographic) bias of FIQA techniques per se, but the performance differentials observed in face recognition systems when used jointly with FIQA approaches. \vspace{2mm} 


%

\section{Related Work}\label{sec:related-work}

A considerable amount of FIQA methods have been presented in the literature over the years following several key ideas. 
One such idea is to extract pseudo-quality ground-truth labels on a closed set of images, which can be used to train a quality estimation network. An example, of a technique from this group is {FaceQnet}~\cite{faceqnet}. FaceQnet uses third-party software to determine the highest quality images of individuals, where the pseudo-quality score is then calculated as the similarity of embeddings of a given image and the predetermined highest quality image of the individual. {PCNet} \cite{pcnet} includes a slightly different labeling approach, where all image pairs are assigned their embedding similarity as the ground truth label and the quality-estimation network is then trained on image pairs. {SDD-FIQA}~\cite{sdd-fiqa} extends on these concepts and incorporates non-mated image pairs into the quality label generation process. 
An emerging idea adopted by several recent methods is to include the quality-estimation process into the training of FR models. The {Probabilistic Face Embeddings~(PFE)}~\cite{pfe} of Shi and Jain trains a FR model to predict the mean and variance vectors for any input, where the mean represents the actual embedding, while the variance can be seen as the uncertainty of the embedding that can be interpreted as the quality of the input image. Similarly, {MagFace} presented by Meng~\textit{et al.} in \cite{magface} introduces a new quality-aware loss capable of predicting the quality of the input sample from the norm of the predicted embedding. One of the most established ideas is to use information directly from the biometric sample or given FR model~\cite{oldufiqa1, oldufiqa2, ser-fiq} to estimate quality. As most methods using this idea deal predominantly with visual quality, their performance is commonly not competitive compared to state-of-the-art FIQA methods. However, two modern approach with competitive results have been proposed recently, i.e., {SER-FIQ}~\cite{ser-fiq} and {FaceQAN}~\cite{babnikICPR2022}. The first relies on measuring differences in embeddings produced by varying the dropout, while the latter relies on adversarial noise.

While the progress made in face-image quality assessment has been impressive, most of the existing work is focused on improved performance. Comprehensive studies exploring the behavior of the models across demographic groups and their fairness towards subjects of a specific sex or race are largely missing from the literature - with the exception of\cite{fiqabias}. We, therefore, address this gap in this paper and present an analysis of the bias of existing FIQA techniques when used jointly with a selected FR model.

\section{Methodology}\label{sec:methods}

The goal of this paper is to examine the differences in the performance and bias of different groups of (face) quality assessment methods. 
In the following section, we present the methodology used for the study and discuss the dataset and models used in the experiments. 

\subsection{Dataset}\label{sec:methods:dataset}

The evaluation was performed using the Balanced Faces in-the Wild (BFW) dataset~\cite{bfw}, which represents a subset of the VGGFace2 dataset. The BFW dataset contains $20000$ face images of $800$ individuals corresponding to classes from two different demographic groups: Sex and Race. For the first group, images are divided into male and female, and for the second group, the images are divided into White, Black, Asian, and Indian\footnote{Note that we intentionally use the terms \textit{sex} and \textit{race} in this work. While \textit{gender} and \textit{sex} have been used interchangeably in the biometric
literature, we follow \cite{TIFS_PrivacySurveyb}, where gender is considered a social or cultural construct, while sex is considered to describe
biological characteristics. The terms race and ethnicity have also
been used interchangeably in the literature, and an exact definition of
these two terms appears to be a subject of debate.}. The entire dataset is balanced by both sex and race, allowing for a fair assessment of the demographic-specific biases inherent in quality assessment methods. Although the dataset was not created specifically for quality assessment, it contains images of varying quality, such as non-frontal or low-resolution images and images with some degree of occlusion. Additionally, the dataset also contains a list of genuine and imposter pairs that are needed for verification experiments to evaluate the biases and performance of face image quality assessment methods.

\subsection{Face Recognition Systems}\label{sec:methods:frsystem}

One of the main goals of our analysis is to explore differences between the results of different FR models. To this end, we use three popular open-source models: {ArcFace}\footnote{\url{https://github.com/deepinsight/insightface}} \cite{arcface}, {VGGFace2}\footnote{\url{https://www.robots.ox.ac.uk/~albanie/pytorch-models.html}} \cite{vggface2} and {FaceNet}\footnote{\url{https://github.com/timesler/facenet-pytorch}} \cite{facenet}. The models differ significantly: ArcFace uses a ResNet100 backbone and an angular-margin loss and  is trained on the MS1MV3 dataset, VGGFace2 uses a SE-ResNet50 backbone and a soft-max loss, and FaceNet uses an Inception-ResNet50 backbone and a triplet loss. Both the VGGFace2 and FaceNet models are trained on the VGGFace2 dataset, used also to construct the BFW dataset, which we use for evaluation. In the evaluation, we therefore study bias with independent test data for ArcFace and data that (partially) overlaps with the training data for VGGFace2 and FaceNet. 

For each of the three FR models, the images are preprocessed as described in the corresponding paper. The embeddings are extracted from the last layer of each model and the cosine similarity is used to generate comparison scores for the verification experiments. 

\subsection{Evaluation Criteria}\label{sec:methods:evaluation}


Following established literature~\cite{cr-fiqa, lightqnet, magface, surveypaper}, we report the performance of the FIQA methods through the use of Error-Versus-Reject Characteristic (ERC) curves, which measure the False Non Match Rate~(FNMR) at a predefined value of False Match Rate~(FMR), typically $0.001$, while increasing the number of rejected low quality images. Additionally, the {Area Under the Curve}~{(AUC)} is computed at different image drop (or reject) rates. When interpreting results, the focus is typically on the lower drop rates, where images of lower quality are rejected.

To assess demographic-specific performance and examine the biases of the FIQA methods when a particular FR model is used, we compare the AUC values of the ERC plots corresponding to different demographic groups. To create the ERC plot for a given  group, we perform demographic-specific verification experiments, in which the images from mated and non-mated pairs all originate from within the same group. We create $100,000$ mated and $300,000$ non-mated image pairs for each demographic group to capture as much within-group variation as possible  and guarantee reliable results. Because we are interested in exploring bias and thus relative performance between groups, we need to normalize the results to allow comparisons between groups. For this reason, the values of the ERC curve from which the AUC is calculated are normalized so that the FNMR value at a drop rate of $0\%$ is equal to $1$, that is, all FNMR values of a given ERC curve are divided by the FNMR value at a drop rate of $0\%$. We denote the normalized variant of the AUC by $AUC_N$.

\section{Presentation of Used Approaches}\label{sec:approaches}


For our research, we use three different groups of methods, that can be used for FIQA. The first group are general purpose {Image Quality Assessment}~{(IQA)}~\cite{brisque, niqe, rankiqa, ssim, fsim} methods, which are different from the other two groups because they are not specifically designed to work with face images , but rather with arbitrary images. The second group of methods are {Supervised Face Image Quality Assessment}~{(sFIQA)}~\cite{magface, sdd-fiqa, cr-fiqa, faceqnet, lightqnet, pcnet} methods, which usually obtain the quality score using a pre-trained quality estimation model. The last group of methods are {Unsupervised Face Image Quality Assessment}~{(uFIQA)}~\cite{ser-fiq, babnikICPR2022} methods that rely only on the information available in the image and a given FR model. In the following sections, we present all three groups and the chosen methods in detail.

\subsection{Image Quality Assessment Methods}\label{sec:approaches:gpiqa}

We use three so-called \textit{no-reference} IQA techniques, i.e.,
{BRISQUE}~\cite{brisque}, {NIQE}~\cite{niqe} and {RankIQA}~\cite{rankiqa}. These techniques are applicable to any input image and unlike other alternatives from the literature~\cite{iqasurvey} require no high quality reference when computing the quality scores.

\textbf{BRISQUE.} The Blind/Referenceless Image Spatial Quality Evaluator presented by Mittal~\textit{et al.} \cite{brisque} tries to estimate the quality characteristics of a given sample using Mean Subtracted Contrast Normalized~(MSCN) coefficients, which for a pristine image exhibit a Gaussian-like shape. For this reason, an Asymmetric Generalized Gaussian Model (AGGM) distribution is used to estimate the MSCN coefficients and a Support Vector Machine~(SVM) is utilized to calculate the final image quality from $32$ extracted features.

\textbf{NIQE.} The Natural Image Quality Evaluator, presented by Mittal~\textit{et al.} \cite{niqe}, is based on quality-aware statistical features of natural scenes obtained from a corpus of natural images. A multivariate Gaussian model is used to fit the coefficients obtained from the corpus. The final quality is calculated as the distance between the model obtained from the natural image corpus and the given image.

\textbf{RankIQA.} This no-reference method of Liu~\textit{et al.} \cite{rankiqa} learns to predict quality from rankings. The rankings are generated using a Siamese network trained to rank images by quality on synthetically generated image sets. Knowledge transfer is then used to train a classical CNN network based on the Siamese model to predict the quality of a given sample.

\subsection{Supervised Face Image Quality Assessment Methods}\label{sec:approaches:sfiqa}

Supervised FIQA methods~\cite{sdd-fiqa, magface, lightqnet, surveypaper} are the most widely used methods in the literature. They usually rely on  pseudo ground-truth quality labels, based on which a quality estimation network is trained. 
We select three widely used state-of-the-art sFIQA methods for our analysis, namely, {SDD-FIQA}~\cite{sdd-fiqa}, {MagFace} \cite{magface}, and CR-FIQA \cite{cr-fiqa}. 

\textbf{SDD-FIQA.} The SSD-FIQA method described by Ou \textit{et al.} \cite{sdd-fiqa}  introduces an advanced unsupervised approach to computing pseudo-quality labels that considers both mated and non-mated image pairs. For a given sample, the quality is computed using the Wasserstein distance between the distributions of mated and non-mated similarity scores by randomly sampling images from a background dataset. The final score is obtained by averaging the partial scores over several runs.

\textbf{MagFace.} The method presented by Meng \textit{et al.} \cite{magface}, called MagFace, generates  both an embedding and a quality score for a given sample by using an extended version of the ArcFace~\cite{arcface} loss. The proposed loss is able to discriminate well between samples of different quality by pushing apart images of different quality. The embeddings generated by a model trained with the new loss can be used to automatically obtain a quality score by measuring their magnitude.

\textbf{CR-FIQA.} The basis for this method of Boutros~\textit{et al.} \cite{cr-fiqa} is the so-called Certainty Ratio~(CR), which is defined in a classification setting when neural networks are trained with a variant of angular-based loss, such as ArcFace. Formally, CR is defined as the ratio between the angular similarities of the face sample and its true class center and the nearest negative class center. A ResNet network is trained on a classification task using a loss composed of the ArcFace and the Certainty Ratio terms. The trained network is then used to predict the quality of a face image.

\subsection{Unsupervised Face Image Quality Assessment Methods}\label{sec:approaches:ufiqa}


A limited number of unsupervised FIQA techniques capable of ensuring state-of-the-art performance has so far been presented in the literature. 
Two such methods are selected for the analysis in this work, i.e.,  {SER-FIQ}~\cite{ser-fiq} and {FaceQAN}~\cite{babnikICPR2022}. Both of these methods have been shown to perform well over a number of FR models and datasets. 

\textbf{SER-FIQ.} Modern FR architectures rely on dropout as a form of regularization when training CNN-based models. The SER-FIQ method, proposed by Terhöerst~\textit{et al.} in \cite{ser-fiq}, uses the dropout layers to measure the quality of a given face image sample. Specifically, for a given sample, a number of different embeddings are created using different sub-network layouts generated by harnessing the dropout layer. The quality is then calculated by measuring the pairwise distances between the constructed features.

\textbf{FaceQAN.} Adversarial approaches are often used to create adversarial examples that can deceive a FR model. The method proposed by Babnik~\textit{et al.} \cite{babnikICPR2022} measures the difficulty of creating adversarial examples in conjunction with a symmetry-estimation process, which incorporates additional information about the facial pose into the quality estimation procedure. The quality score is calculated from statistics derived from the similarity between adversarial and input sample embeddings multiplied by the symmetry score.

\begin{table}[!ht]
    \centering
    \resizebox{.99\columnwidth}{!}{%
    \begin{tabular}{|c|| c | c | c |}
        \hline
        \textbf{Model} & ArcFace & VGGFace2 & FaceNet \\
        \hline
        \textbf{TAR@FAR(1e-4)[\%]} & $95.3$ & $86.2$ & $75.3$  \\
        \hline
    \end{tabular}
    }
    \caption{Verification performance on the IJB-C dataset.}
    \label{tab:fr_perf}
\end{table}

\begin{figure}[ht!]
    \centering
    \setlength\extrarowheight{1.9pt}
    \includegraphics[width=0.4\linewidth, trim = 2mm 21mm 3mm 2mm,clip ]{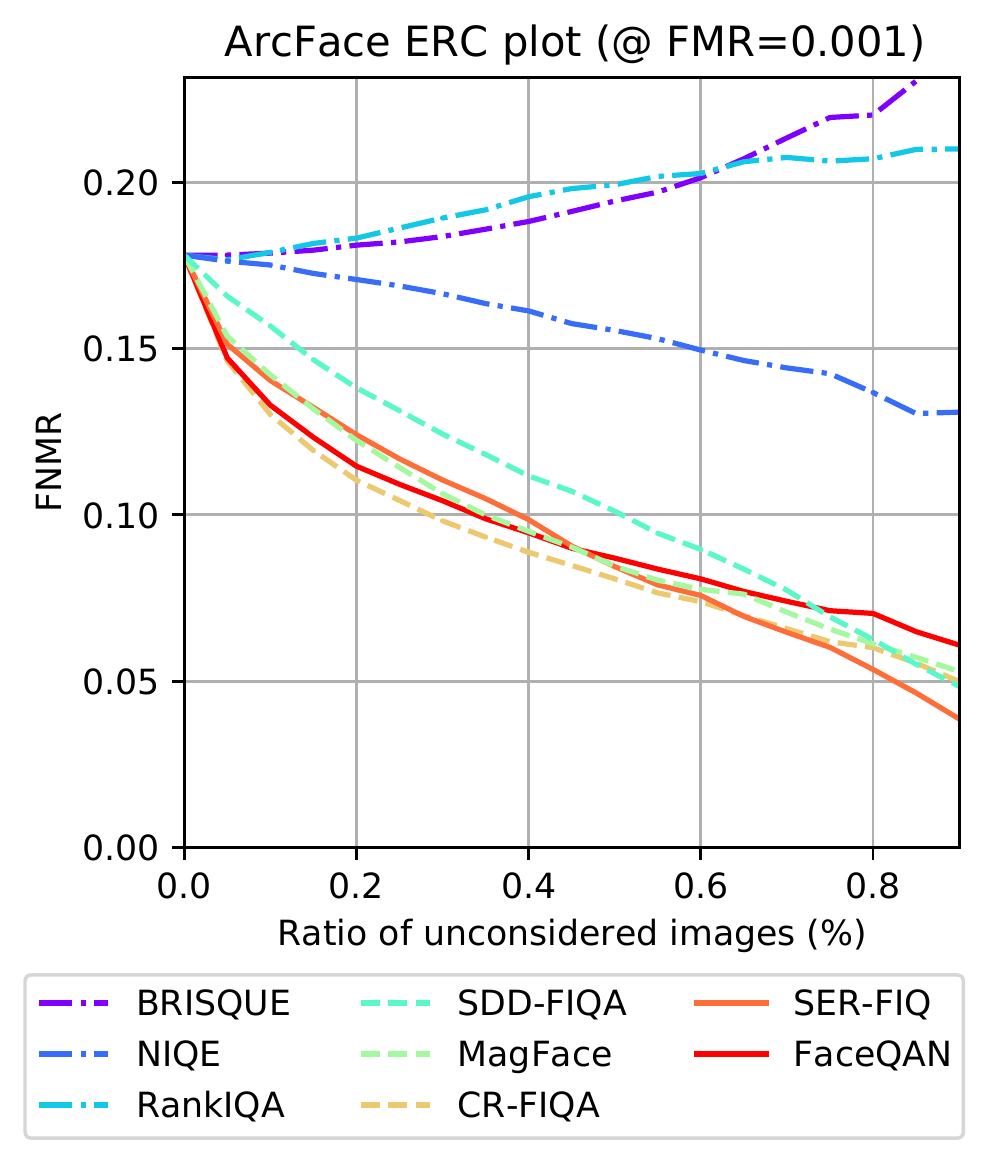}
    \raisebox{1.2\height}{
        \resizebox{0.56\linewidth}{!}{%
            \begin{tabular}{c| c c c c}
            
            \hline
            
              \multirow{2}{*}{\textbf{Methods}} & \multicolumn{4}{c}{\textbf{Drop Rate}} \\ 
                & $\mathbf{10\%}$ & $\mathbf{20\%}$ & $\mathbf{40\%}$ & $\mathbf{80\%}$ \\  
            
            \hline

\textbf{BRISQUE} & $0.0178$ & $0.0358$ & $0.0726$ & $0.1540$\\
\textbf{NIQE} & \cellcolor{red!10}$0.0176$ & \cellcolor{red!10}$0.0349$ & \cellcolor{red!10}$0.0682$ & \cellcolor{red!10}$0.1280$\\
\textbf{RankIQA} & $0.0178$ & $0.0359$ & $0.0737$ & $0.1549$\\

\hline

\textbf{SDD-FIQA} & $0.0167$ & $0.0314$ & $0.0563$ & $0.0918$\\
\textbf{MagFace} & $0.0157$ & $0.0289$ & $0.0504$ & $0.0815$\\
\textbf{CR-FIQA} & \cellcolor{green!10}$\mathbf{0.0150}$ & \cellcolor{green!10}$\mathbf{0.0270}$ & \cellcolor{green!10}$\mathbf{0.0468}$ & \cellcolor{green!10}$\mathbf{0.0762}$\\

\hline

\textbf{SER-FIQ} & $0.0155$ & $0.0288$ & $0.0509$ & $0.0809$\\
\textbf{FaceQAN} & \cellcolor{blue!10}$0.0151$ & \cellcolor{blue!10}$0.0275$ & \cellcolor{blue!10}$0.0483$ & \cellcolor{blue!10}$0.0806$\\

                \hline
            \end{tabular}
        }
    }
    
    \includegraphics[width=0.4\linewidth, trim = 2mm 21mm 3mm 2mm,clip ]{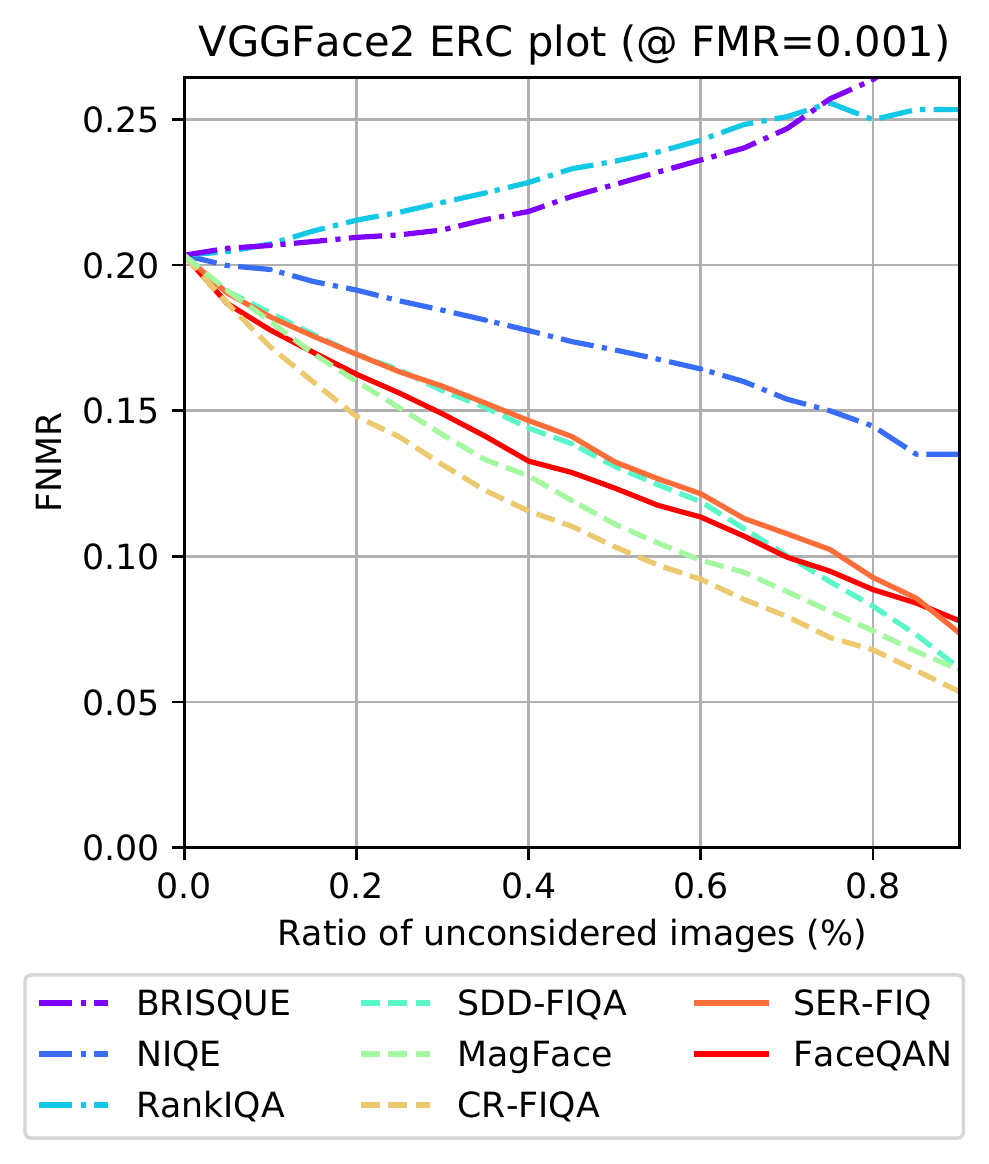}
    \raisebox{1.2\height}{
        \resizebox{0.56\linewidth}{!}{%
            \begin{tabular}{c|c c c c}
            
            \hline
              \multirow{2}{*}{\textbf{Methods}} & \multicolumn{4}{c}{\textbf{Drop Rate}} \\ 
                & $\mathbf{10\%}$ & $\mathbf{20\%}$ & $\mathbf{40\%}$ & $\mathbf{80\%}$ \\  
              
             \hline

\textbf{BRISQUE} & $0.0205$ & $0.0414$ & $0.0840$ & $0.1792$\\
\textbf{NIQE} & \cellcolor{red!10}$0.0200$ & \cellcolor{red!10}$0.0395$ & \cellcolor{red!10}$0.0764$ & \cellcolor{red!10}$0.1415$\\
\textbf{RankIQA} & $0.0205$ & $0.0416$ & $0.0860$ & $0.1832$\\

\hline

\textbf{SDD-FIQA} & $0.0192$ & $0.0369$ & $0.0683$ & $0.1146$\\
\textbf{MagFace} & $0.0192$ & $0.0362$ & $0.0646$ & $0.1045$\\
\textbf{CR-FIQA} & \cellcolor{green!10}$\mathbf{0.0187}$ & \cellcolor{green!10}$\mathbf{0.0347}$ & \cellcolor{green!10}$\mathbf{0.0611}$ & \cellcolor{green!10}$\mathbf{0.0976}$\\

\hline

\textbf{SER-FIQ} & $0.0192$ & $0.0367$ & $0.0683$ & $0.1166$\\
\textbf{FaceQAN} & \cellcolor{blue!10}$0.0189$ & \cellcolor{blue!10}$0.0359$ & \cellcolor{blue!10}$0.0656$ & \cellcolor{blue!10}$0.1103$\\

                \hline
            \end{tabular}
        }
    }
    
    \includegraphics[width=0.4\linewidth, trim = 2mm 21mm 3mm 2mm,clip ]{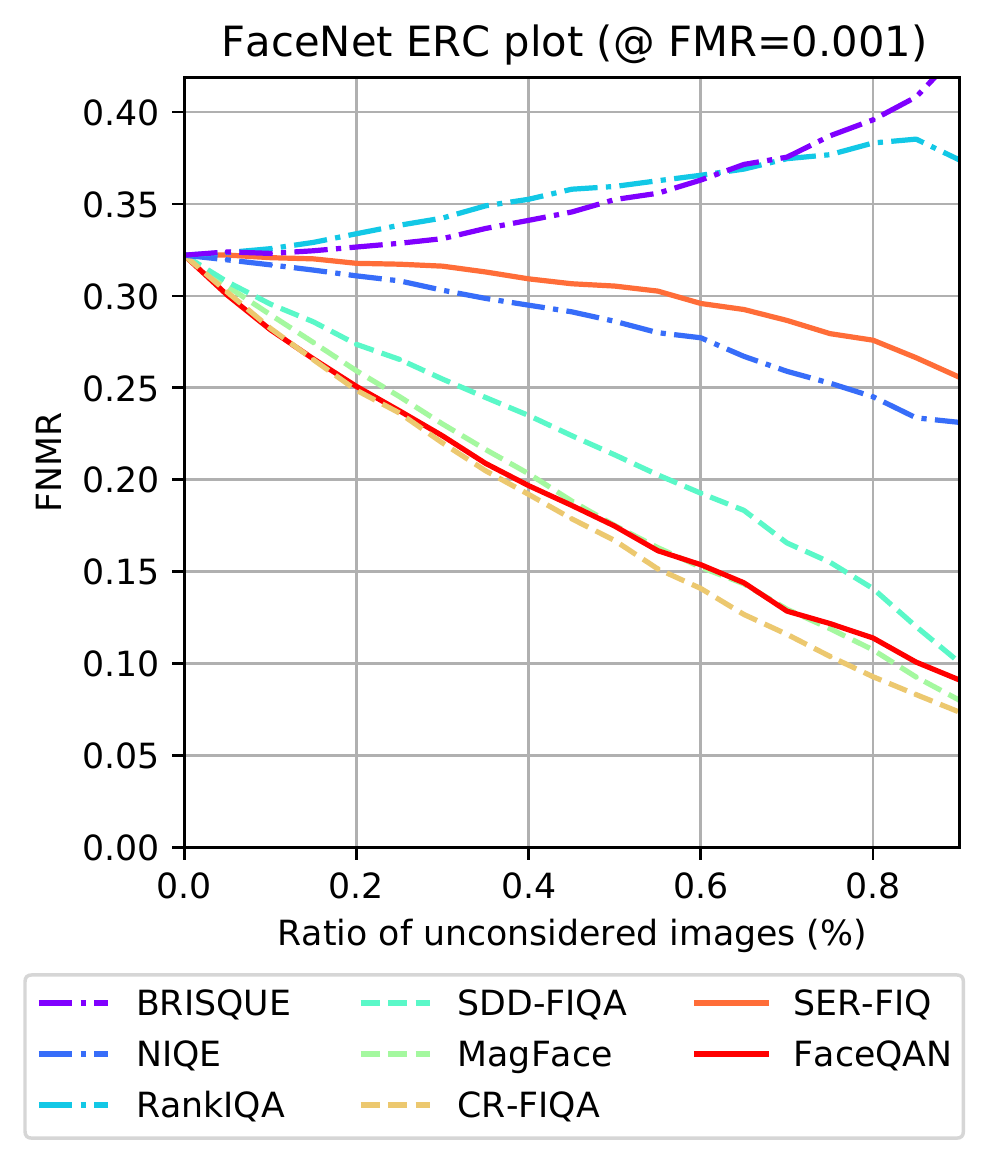}
    \raisebox{1.2\height}{
        \resizebox{0.56\linewidth}{!}{%
            \begin{tabular}{c|c c c c}
            
            \hline
            
              \multirow{2}{*}{\textbf{Methods}} & \multicolumn{4}{c}{\textbf{Drop Rate}} \\ 
                & $\mathbf{10\%}$ & $\mathbf{20\%}$ & $\mathbf{40\%}$ & $\mathbf{80\%}$ \\
              
            \hline

\textbf{BRISQUE} & $0.0323$ & $0.0648$ & $0.1313$ & $0.2773$\\
\textbf{NIQE} & \cellcolor{red!10}$0.0320$ & \cellcolor{red!10}$0.0634$ & \cellcolor{red!10}$0.1240$ & \cellcolor{red!10}$0.2332$\\
\textbf{RankIQA} & $0.0324$ & $0.0653$ & $0.1340$ & $0.2807$\\

\hline

\textbf{SDD-FIQA} & $0.0308$ & $0.0593$ & $0.1103$ & $0.1866$\\
\textbf{MagFace} & $0.0306$ & $0.0580$ & $0.1042$ & $0.1655$\\
\textbf{CR-FIQA} & \cellcolor{green!10}$0.0302$ & \cellcolor{green!10}$0.0568$ & \cellcolor{green!10}$\mathbf{0.1008}$ & \cellcolor{green!10}$\mathbf{0.1572}$\\

\hline

\textbf{SER-FIQ} & $0.0322$ & $0.0642$ & $0.1271$ & $0.2452$\\
\textbf{FaceQAN} & \cellcolor{blue!10}$\mathbf{0.0301}$ & \cellcolor{blue!10}$\mathbf{0.0567}$ & \cellcolor{blue!10}$0.1014$ & \cellcolor{blue!10}$0.1627$\\

                \hline
            \end{tabular}
        }
    }
    
    \includegraphics[width=0.5\linewidth, trim = 0mm 0mm 0mm 0mm,clip ]{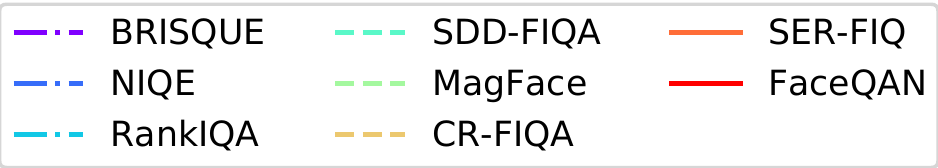}\vspace{2mm}
    \resizebox{0.75\linewidth}{!}{%
    {\scriptsize
    \centering
    \hspace{2mm}
    \begin{tabular}{rr l r l }
        &\cellcolor{red!10} & -- best IQA approach, & \cellcolor{green!10} & -- best sFIQA approach, \\
        &\cellcolor{blue!10} & -- best uFIQA approach, & \textbf{B} & -- best overall.
    \end{tabular}
    }
    }
    \caption{\textbf{Results of the performance evaluation.} The performance is reported using ERC curve plots~(left) and AUC scores ($\downarrow$) at different drop rates~(right). \\
       \vspace{-4mm}  }
    \label{fig:performance_results}
\end{figure}

\section{Experiments and results}\label{sec:experiments-results}

In the following section we present the results of our experiments conducted to investigate: $(i)$ the performance and (demographic) bias of the FIQA methods, $(ii)$ the performance differences between IQA, sFIQA and uFIQA techniques, and $(iii)$ the impact of FR models on the observed results. 

\subsection{Face Image Quality Assessment Performance.} 

We first evaluate the performance of individual FIQA methods using standard evaluation methodology to benchmark their performance with different FR models. To give a better overview of the standings between the FR models, we present verification performances on the IJB-C dataset in Table~\ref{tab:fr_perf}, whereas the FIQA results in the form of ERC curve plots and AUC scores for different drop rates are shown in Fig.~\ref{fig:performance_results}.

\textbf{Comparison of IQA Methods.} Overall, the results of all tested IQA methods tell the same story regardless of the  FR model used. NIQE is by far the best performing method, while BRISQUE and RankIQA generally perform worse. Looking at the ERC plots, we see that NIQE provides quality scores that lower the FNMR, while both BRISQUE and RankIQA seem to increase the FNMR indicating that they are of limited use as additional sources of information for FR models.

\textbf{Comparison of sFIQA Methods.} All evaluated sFIQA methods appear to be highly effective as a sharp decline in the FNMR can be seen with increasing reject rates. Overall, the CR-FIQA method perform the best with all considered FR models, followed closely by MagFace. While SDD-FIQA is the weakest of the three methods, its performance is still relatively close to both CR-FIQA and MagFace.

\textbf{Comparison of uFIQA Methods.} Similarily to sFIQA, both evaluated uFIQA approaches perform well with all three FR models, with a notable exception of SER-FIQ and the FaceNet model. Both methods are highly competitive on ArcFace and VGGFace, with FaceQAN having a slight advantage over SER-FIQ on both.

\textbf{Overall Comparison.} Considering all $3$ FR models, CR-FIQA performs best overall, followed closely by FaceQAN. Both the sFIQA and uFIQA groups are highly competitive, whereas the IQA methods fall short when it comes to performance, which is expected since these methods were not designed specifically with FR in mind.

\subsection{Bias Analysis}

We explore differences in the demographic-specific AUC scores produced by the evaluated quality assessment techniques in Table~\ref{tab:bias_res_fld}. 
Here, we report  $AUC_N$ scores for the baseline verification experiments as well as scores for each demographic group separately. Additionally, we also calculate the relative difference of the $AUC_N$ for a given demographic group w.r.t. the best performing group -- marked with $(B)$. 

\textbf{Impact of Sex.} 
For the ArcFace model, the results for the  IQA and sFIQA methods show a clear preference for females, whereas no consistent trend can be observed for the uFIQA methods. Nevertheless, the relative differences between men and women are limited (below $10\%$) for all methods, except for MagFace with a relative difference of $13\%$. The results for the other two FR models show smaller performance differential across sex for the majority of tested techniques compared to the ArcFace model, 
suggesting that sex bias is not a major problem for quality assessment methods.
\begin{table*}[t]
    \centering
    \renewcommand{\arraystretch}{1.1}
    \caption{$AUC_N$ scores generated for the bias-related experiments. $R^i_B = (\sfrac{AUC^{i}_N}{min_{\{i\}}AUC^{i}_N}) - 1$}\vspace{-2mm}
    \resizebox{0.8\linewidth}{!}{%
    \begin{tabular}{| c | l | c | l l | l l l l }
        \hline
        \multirow{2}{*}{\textbf{FR model}} & \multirow{2}{*}{\textbf{Method}} & \multirow{2}{*}{$\mathbf{AUC_N(\downarrow)}$} & \multicolumn{2}{c|}{\textbf{Sex--specific} $\mathbf{AUC_N^s(\downarrow)}$ ($R_B^s \%$)} &  \multicolumn{4}{c|}{\textbf{Race--specific} $\mathbf{AUC_N^r(\downarrow)}$ ($R_B^r \%$)} \\
            \hhline{*{3}{|~}*{2}{-}*{4}{|-}}
        & & &  \textbf{Male} & \textbf{Female} & \textbf{White} & \textbf{Black} & \textbf{Asian} & \multicolumn{1}{c|}{\textbf{Indian}} \\
         \hline
         \parbox[t]{2mm}{\multirow{8}{*}{\rotatebox[origin=c]{90}{\textbf{ArcFace}}}} 
         
 & BRISQUE & $1.066$ &  $1.078$ ($\bm{1.8{\%}}$) & \multicolumn{1}{l|}{$1.059$ ($B$)}  & $1.094$ ($\bm{8.3{\%}}$) & $1.067$ ($5.6\%$) & $1.010$ ($B$) & \multicolumn{1}{l|}{$1.044$ ($3.4\%$)} \\
 & NIQE & $0.831$ &  $0.835$ ($\bm{1.2{\%}}$) & \multicolumn{1}{l|}{$0.825$ ($B$)} & $0.843$ ($1.1\%$) & $0.854$ ($2.4\%$) & $0.867$ ($\bm{4.0{\%}}$) & \multicolumn{1}{l|}{$0.834$ ($B$)} \\
 & RankIQA & $1.047$ &  \cellcolor{red!10}$1.079$ ($\bm{4.4{\%}}$) & \multicolumn{1}{l|}{$1.034$ ($B$)}  & $1.061$ ($6.2\%$) & $1.082$ ($8.3\%$) & $0.999$ ($B$) & \multicolumn{1}{l|}{\cellcolor{red!10}$1.108$ ($\bm{10.9{\%}}$)}\\
\hhline{~*{8}{-}}
 & SDD-FIQA & $0.560$ &  $0.573$ ($\bm{1.1{\%}}$) & \multicolumn{1}{l|}{$0.567$ ($B$)}  & $0.513$ ($B$) & $0.602$ ($17.3\%$) & $0.615$ ($\bm{19.9{\%}}$) & \multicolumn{1}{l|}{$0.554$ ($8.0\%$)} \\
 & MagFace & $0.504$ &  \cellcolor{green!10}$0.549$ ($\bm{13.0{\%}}$) & \multicolumn{1}{l|}{$0.486$ ($B$)} & $0.430$ ($B$) & $0.564$ ($\bm{31.2{\%}}$) & $0.523$ ($21.6\%$) & \multicolumn{1}{l|}{$0.518$ ($20.5\%$)} \\
 & CR-FIQA & $0.472$ &  $0.507$ ($\bm{9.5{\%}}$) & \multicolumn{1}{l|}{$0.463$ ($B$)}  & $0.376$ ($B$) & \cellcolor{green!10}$0.527$ ($\bm{40.2{\%}}$) & $0.512$ ($36.2\%$) & \multicolumn{1}{l|}{$0.493$ ($31.1\%$)}\\
\hhline{~*{8}{-}}
 & SER-FIQ & $0.490$ &  $0.488$ ($B$) & \multicolumn{1}{l|}{$0.518$ ($\bm{6.1{\%}}$)} & $0.438$ ($B$) & $0.576$ ($\bm{31.5{\%}}$) & $0.572$ ($30.6\%$) & \multicolumn{1}{l|}{$0.539$ ($23.1\%$)} \\
 & FaceQAN & $0.507$ &  \cellcolor{blue!10}$0.542$ ($\bm{8.8{\%}}$) & \multicolumn{1}{l|}{$0.498$ ($B$)} & $0.432$ ($B$) & $0.556$ ($28.7\%$) & $0.547$ ($26.6\%$) & \multicolumn{1}{l|}{$0.569$ \cellcolor{blue!10}($\bm{31.7{\%}}$)} \\

         \hline\hline
         
         \parbox[t]{2mm}{\multirow{8}{*}{\rotatebox[origin=c]{90}{\textbf{VGGFace2}}}}

 & BRISQUE & $1.088$ &  $1.121$ ($\bm{1.4{\%}}$) & \multicolumn{1}{l|}{$1.106$ ($B$)} & $1.135$ ($\bm{11.4{\%}}$) & $1.049$ ($2.9\%$) & $1.019$ ($B$) & \multicolumn{1}{l|}{$1.090$ ($7.0\%$)} \\
 & NIQE & $0.795$ &  $0.781$ ($B$) & \multicolumn{1}{l|}{\cellcolor{red!10}$0.809$ ($\bm{3.6{\%}}$)} & $0.809$ ($B$) & $0.841$ ($4.0\%$) & $0.855$ ($\bm{5.7{\%}}$) & \multicolumn{1}{l|}{$0.832$ ($2.8\%$)} \\
 & RankIQA & $1.088$ &  $1.083$ ($\bm{1.3{\%}}$) & \multicolumn{1}{l|}{$1.069$ ($B$)} & $1.090$ ($10.1\%$) & $1.054$ ($6.5\%$) & $0.990$ ($B$) & \multicolumn{1}{l|}{\cellcolor{red!10}$1.106$ ($\bm{11.7{\%}}$)} \\
\hhline{~*{8}{-}}
 & SDD-FIQA & $0.614$ &  $0.605$ ($B$) & \multicolumn{1}{l|}{$0.622$ ($\bm{2.8{\%}}$)} & $0.562$ ($B$) & $0.634$ ($12.8\%$) & $0.673$ ($\bm{19.8{\%}}$) & \multicolumn{1}{l|}{$0.597$ ($6.2\%$)} \\
 & MagFace & $0.561$ &  \cellcolor{green!10}$0.576$ ($\bm{4.0{\%}}$) & \multicolumn{1}{l|}{$0.554$ ($B$)} & $0.501$ ($B$) & $0.584$ ($16.6\%$) & $0.601$ ($\bm{20.0{\%}}$) & \multicolumn{1}{l|}{$0.560$ ($11.8\%$)} \\
 & CR-FIQA & $0.522$ &  $0.538$ ($\bm{3.1{\%}}$) & \multicolumn{1}{l|}{$0.522$ ($B$)} & $0.466$ ($B$) & $0.542$ ($16.3\%$) & \cellcolor{green!10}$0.575$ ($\bm{23.4{\%}}$) & \multicolumn{1}{l|}{$0.532$ ($14.2\%$)}\\
\hhline{~*{8}{-}}
 & SER-FIQ & $0.631$ &  $0.632$ ($\bm{0.8{\%}}$) & \multicolumn{1}{l|}{$0.627$ ($B$)} & $0.626$ ($B$) & \cellcolor{blue!10}$0.704$ ($\bm{12.5{\%}}$) & $0.699$ ($11.7\%$) & \multicolumn{1}{l|}{$0.673$ ($7.5\%$)}\\
 & FaceQAN & $0.601$ &  $0.602$ ($B$) & \multicolumn{1}{l|}{\cellcolor{blue!10}$0.632$ ($\bm{5.0{\%}}$)} & $0.598$ ($B$) & $0.605$ ($1.2\%$) & $0.647$ ($\bm{8.2{\%}}$) & \multicolumn{1}{l|}{$0.623$ ($4.2\%$)} \\
         
         \hline\hline
         
         \parbox[t]{2mm}{\multirow{8}{*}{\rotatebox[origin=c]{90}{\textbf{FaceNet}}}}

 & BRISQUE & $1.058$ &  \cellcolor{red!10}$1.141$ ($\bm{7.4{\%}}$) & \multicolumn{1}{l|}{$1.062$ ($B$)} & \cellcolor{red!10}$1.156$ ($\bm{15.6{\%}}$) & $1.108$ ($10.8\%$) & $1.000$ ($B$) & \multicolumn{1}{l|}{$1.123$ ($12.3\%$)} \\
 & NIQE & $0.831$ &  $0.799$ ($B$) & \multicolumn{1}{l|}{$0.851$ ($\bm{6.5{\%}}$)} & $0.786$ ($B$) & $0.844$ ($7.4\%$) & $0.881$ ($\bm{12.1{\%}}$) & \multicolumn{1}{l|}{$0.818$ ($4.1\%$)} \\
 & RankIQA & $1.048$ &  $1.069$ ($\bm{3.3{\%}}$) & \multicolumn{1}{l|}{$1.035$ ($B$)} & $1.126$ ($\bm{13.6{\%}}$) & $1.047$ ($5.7\%$) & $0.991$ ($B$) & \multicolumn{1}{l|}{$1.085$ ($9.5\%$)} \\
\hhline{~*{8}{-}}
 & SDD-FIQA & $0.630$ &  $0.602$ ($B$) & \multicolumn{1}{l|}{\cellcolor{green!10}$0.635$ ($\bm{5.5{\%}}$)} & $0.557$ ($B$) & $0.644$ ($15.6\%$) & $0.701$ ($\bm{25.9{\%}}$) & \multicolumn{1}{l|}{$0.616$ ($10.6\%$)} \\
 & MagFace & $0.554$ &  $0.539$ ($\bm{0.9{\%}}$) & \multicolumn{1}{l|}{$0.534$ ($B$)}  & $0.466$ ($B$) & $0.585$ ($25.5\%$) & $0.620$ ($\bm{33.0{\%}}$) & \multicolumn{1}{l|}{$0.529$ ($13.5\%$)}  \\
 & CR-FIQA & $0.524$ &  $0.514$ ($B$) & \multicolumn{1}{l|}{$0.520$ ($\bm{1.2{\%}}$)} & $0.437$ ($B$) & $0.551$ ($26.1\%$) & \cellcolor{green!10}$0.597$ ($\bm{36.6{\%}}$) & \multicolumn{1}{l|}{$0.528$ ($20.8\%$)} \\
\hhline{~*{8}{-}}
 & SER-FIQ & $0.882$ &  $0.857$ ($B$) & \multicolumn{1}{l|}{\cellcolor{blue!10}$0.881$ ($\bm{2.8{\%}}$)} & $0.900$ ($5.4\%$) & $0.994$ ($\bm{16.4{\%}}$) & $0.854$ ($B$) & \multicolumn{1}{l|}{$0.884$ ($3.5\%$)} \\
 & FaceQAN & $0.550$ &  $0.555$ ($B$) & \multicolumn{1}{l|}{$0.556$ ($\bm{0.2{\%}}$)} & $0.449$ ($B$) & $0.574$ ($27.8\%$) & \cellcolor{blue!10}$0.617$ ($\bm{37.4{\%}}$) & \multicolumn{1}{l|}{$0.577$ ($28.5\%$)}  \\
         
         \hline
         
    \end{tabular}
    }
    
    {\footnotesize
        \centering
        \begin{tabular}{r l r l r l}
            \cellcolor{red!10} & -- weakest IQA performance , & \cellcolor{green!10} & -- weakest sFIQA performance, & \cellcolor{blue!10} & -- weakest uFIQA performance.
        \end{tabular}
    }
    \label{tab:bias_res_fld}\vspace{-2mm}
\end{table*}

\textbf{Impact of Race.} 
As seen from the results in Table~\ref{tab:bias_res_fld}, the relative differences in performance between the race groups are much larger than for the two sex categories. As expected, the bias of IQA methods appears to be less pronounced compared to the sFIQA and uFIQA methods, as they do not favour any particular race overall. Two out of three methods from this group perform best for Asian subjects and one for Indian subjects. The performance differentials are at most around $15\%$ (BRISQUE) when comparing the best and worst performing groups.
The results for the sFIQA methods are consistent. All tested methods perform best for White subjects with all three FR models and worst on images from the Black or Asian categories -- depending on the FR model used. In the case of VGGFace2 and FaceNet, Asian individuals appear to be the least favoured, while for ArcFace the results for the methods are mixed. Overall there appears to be a strong bias towards white people in all FR-FIQA model combinations. 
Similarly, the uFIQA methods show a strong preference towards White subjects, with the exception of SER-FIQ when using FaceNet. Both uFIQA methods perform worst with subjects from the Black and Asian group, with the exception of FaceQAN when using ArcFace, which performs worst with Indian subjects.
The effect of overlapping data on both the VGGFace2 and FaceNet model seems to cause mixed results, while for VGGFace2 the bias is overall lower than that of the ArcFace model, the bias for FaceNet seems comparable or rather larger than that of the ArcFace model.
Another interesting observation is that the performance of individual methods appears to be related to racial bias, specifically how strong the method's preference is for white people. The better the performance of the method, the more it appears to favour White subjects over other races. The main methods that exhibit this behavior are the highest performing CR-FIQA and FaceQAN. Another example where this can be seen is SER-FIQ on FaceNet, where the method seems to weaken in its performance but shows a weaker preference for White individuals, and NIQE, which performs best of all IQA methods but also seems to favour White subjects more.

\section{Conclusion}\label{sec:conclusion}

In this paper, we presented an analysis of the performance and biases of different quality assessment methods separated into three groups: Image Quality Assessment, Supervised Face Image Quality Assessment, and Unsupervised Face Image Quality Assessment, using three different face recognition models. The results show that supervised and unsupervised face image quality assessment methods are highly competitive across all face recognition models, with CR-FIQA coming out on top in most cases. General image quality assessment methods, on the other hand, perform worse because they assess visual quality rather than biometric utility of the samples. The bias experiments showed stronger results for supervised and unsupervised methods with respect to White subjects, with the worst results for individuals from the Black and Asian group. In addition, methods that exhibited greater bias appeared to perform better overall, leading to the assumption that the observed bias is related to a considerable extent to the underlying face recognition model. This observation opens up possibilities for future research, as debiasing schemes would have to consider quality assessment and face recognition in a joint setting to be able to effectively reduce performance differentials across different demographic groups. 

\bibliographystyle{IEEEtran}
\bibliography{bibliography}

\begin{thebibliography}{10}
\providecommand{\url}[1]{#1}
\csname url@samestyle\endcsname
\providecommand{\newblock}{\relax}
\providecommand{\bibinfo}[2]{#2}
\providecommand{\BIBentrySTDinterwordspacing}{\spaceskip=0pt\relax}
\providecommand{\BIBentryALTinterwordstretchfactor}{4}
\providecommand{\BIBentryALTinterwordspacing}{\spaceskip=\fontdimen2\font plus
\BIBentryALTinterwordstretchfactor\fontdimen3\font minus
  \fontdimen4\font\relax}
\providecommand{\BIBforeignlanguage}[2]{{%
\expandafter\ifx\csname l@#1\endcsname\relax
\typeout{** WARNING: IEEEtran.bst: No hyphenation pattern has been}%
\typeout{** loaded for the language `#1'. Using the pattern for}%
\typeout{** the default language instead.}%
\else
\language=\csname l@#1\endcsname
\fi
#2}}
\providecommand{\BIBdecl}{\relax}
\BIBdecl

\bibitem{grm2018strengths}
K.~Grm, V.~{\v{S}}truc, A.~Artiges, M.~Caron, and H.~K. Ekenel, ``Strengths and
  weaknesses of deep learning models for face recognition against image
  degradations,'' \emph{IET Biometrics}, vol.~7, no.~1, pp. 81--89, 2018.

\bibitem{frbias1}
A.~Kortylewski, B.~Egger, A.~Schneider, T.~Gerig, A.~Morel-Forster, and
  T.~Vetter, ``Analyzing and reducing the damage of dataset bias to face
  recognition with synthetic data,'' in \emph{CVPR-W}, 2019.

\bibitem{frbias2}
C.~Huang, Y.~Li, C.~C. Loy, and X.~Tang, ``Deep imbalanced learning for face
  recognition and attribute prediction,'' \emph{IEEE Transactions on Pattern
  Analysis and Machine Intelligence}, vol.~42, no.~11, pp. 2781--2794, 2019.

\bibitem{frbias3}
J.~P. Robinson, G.~Livitz, Y.~Henon, C.~Qin, Y.~Fu, and S.~Timoner, ``Face
  recognition: too bias, or not too bias?'' in \emph{CVPR-W}, 2020.

\bibitem{frbias4}
J.~G. Cavazos, P.~J. Phillips, C.~D. Castillo, and A.~J. O’Toole, ``Accuracy
  comparison across face recognition algorithms: Where are we on measuring race
  bias?'' \emph{IEEE Transactions on Biometrics, Behavior, and Identity
  Science}, vol.~3, no.~1, pp. 101--111, 2020.

\bibitem{fiqabias}
P.~Terh{\"o}rst, J.~N. Kolf, N.~Damer, F.~Kirchbuchner, and A.~Kuijper, ``Face
  quality estimation and its correlation to demographic and non-demographic
  bias in face recognition,'' in \emph{IJCB}, 2020, pp. 1--11.

\bibitem{faceqnet}
J.~Hernandez-Ortega, J.~Galbally, J.~Fierrez, R.~Haraksim, and L.~Beslay,
  ``Faceqnet: Quality assessment for face recognition based on deep learning,''
  in \emph{ICB}, 2019, pp. 1--8.

\bibitem{pcnet}
W.~Xie, J.~Byrne, and A.~Zisserman, ``Inducing predictive uncertainty
  estimation for face verification,'' in \emph{BMVC}, 2020.

\bibitem{sdd-fiqa}
F.-Z. Ou, X.~Chen, R.~Zhang, Y.~Huang, S.~Li, J.~Li, Y.~Li, L.~Cao, and Y.-G.
  Wang, ``{SDD-FIQA: Unsupervised face image quality assessment with similarity
  distribution distance},'' in \emph{CVPR}, 2021, pp. 7670--7679.

\bibitem{pfe}
Y.~Shi and A.~K. Jain, ``Probabilistic face embeddings,'' in \emph{ICCV}, 2019.

\bibitem{magface}
Q.~Meng, S.~Zhao, Z.~Huang, and F.~Zhou, ``Magface: A universal representation
  for face recognition and quality assessment,'' in \emph{CVPR}, 2021, pp.
  14\,225--14\,234.

\bibitem{oldufiqa1}
X.~Gao, S.~Z. Li, R.~Liu, and P.~Zhang, ``Standardization of face image sample
  quality,'' in \emph{ICB}, 2007, pp. 242--251.

\bibitem{oldufiqa2}
A.~Abaza, M.~A. Harrison, and T.~Bourlai, ``Quality metrics for practical face
  recognition,'' in \emph{ICPR}, 2012, pp. 3103--3107.

\bibitem{ser-fiq}
P.~Terh{\"o}rst, J.~N. Kolf, N.~Damer, F.~Kirchbuchner, and A.~Kuijper,
  ``Ser-fiq: Unsupervised estimation of face image quality based on stochastic
  embedding robustness,'' in \emph{CVPR}, 2020, pp. 5651--5660.

\bibitem{babnikICPR2022}
{\v{Z}}.~Babnik, P.~Peer, and V.~Štruc, ``Faceqan: Face image quality
  assessment through adversarial noise exploration,'' in \emph{ICPR}, 2022.

\bibitem{bfw}
J.~P. Robinson, G.~Livitz, Y.~Henon, C.~Qin, Y.~Fu, and S.~Timoner, ``Face
  recognition: Too bias, or not too bias?'' in \emph{CVPR-W}, 2020.

\bibitem{TIFS_PrivacySurveyb}
B.~Meden, P.~Rot, P.~Terh{\"o}rst, N.~Damer, A.~Kuijper, J.~W. Scheirer,
  A.~Ross, P.~Peer, and V.~Štruc, ``Privacy-enhancing face biometrics: A
  comprehensive survey,'' \emph{IEEE Transactions on Information Forensics and
  Security}, vol.~16, pp. 4147--4183, 2021.

\bibitem{arcface}
J.~Deng, J.~Guo, N.~Xue, and S.~Zafeiriou, ``Arcface: Additive angular margin
  loss for deep face recognition,'' in \emph{CVPR}, 2019, pp. 4690--4699.

\bibitem{vggface2}
Q.~Cao, L.~Shen, W.~Xie, O.~M. Parkhi, and A.~Zisserman, ``Vggface2: A dataset
  for recognising faces across pose and age,'' in \emph{FG}, 2018.

\bibitem{facenet}
F.~Schroff, D.~Kalenichenko, and J.~Philbin, ``Facenet: A unified embedding for
  face recognition and clustering,'' in \emph{CVPR}, 2015, pp. 815--823.

\bibitem{cr-fiqa}
F.~Boutros, M.~Fang, M.~Klemt, B.~Fu, and N.~Damer, ``Cr-fiqa: Face image
  quality assessment by learning sample relative classifiability,'' \emph{arXiv
  preprint arXiv:2112.06592}, 2021.

\bibitem{lightqnet}
K.~Chen, T.~Yi, and Q.~Lv, ``Lightqnet: Lightweight deep face quality
  assessment for risk-controlled face recognition,'' \emph{IEEE Signal
  Processing Letters}, vol.~28, pp. 1878--1882, 2021.

\bibitem{surveypaper}
T.~Schlett, C.~Rathgeb, O.~Henniger, J.~Galbally, J.~Fierrez, and C.~Busch,
  ``Face image quality assessment: A literature survey,'' \emph{ACM Computing
  Surveys}, 2022.

\bibitem{brisque}
A.~Mittal, A.~K. Moorthy, and A.~C. Bovik, ``Blind/referenceless image spatial
  quality evaluator,'' in \emph{ASILOMAR}, 2011, pp. 723--727.

\bibitem{niqe}
A.~Mittal, R.~Soundararajan, and A.~C. Bovik, ``Making a “completely blind”
  image quality analyzer,'' \emph{IEEE Signal Processing Letters}, 2012.

\bibitem{rankiqa}
X.~Liu, J.~Van De~Weijer, and A.~D. Bagdanov, ``Rankiqa: Learning from rankings
  for no-reference image quality assessment,'' in \emph{ICCV}, 2017.

\bibitem{ssim}
Z.~Wang, A.~C. Bovik, H.~R. Sheikh, and E.~P. Simoncelli, ``Image quality
  assessment: from error visibility to structural similarity,'' \emph{IEEE
  Transactions on Image Processing}, vol.~13, no.~4, pp. 600--612, 2004.

\bibitem{fsim}
L.~Zhang, L.~Zhang, X.~Mou, and D.~Zhang, ``Fsim: A feature similarity index
  for image quality assessment,'' \emph{IEEE Transactions on Image Processing},
  vol.~20, no.~8, pp. 2378--2386, 2011.

\bibitem{iqasurvey}
G.~Zhai and X.~Min, ``Perceptual image quality assessment: a survey,''
  \emph{Science China Information Sciences}, vol.~63, no.~11, pp. 1--52, 2020.

\end{thebibliography}


\begin{thebibliography}{00}
\bibitem{b1} G. Eason, B. Noble, and I. N. Sneddon, ``On certain integrals of Lipschitz-Hankel type involving products of Bessel functions,'' Phil. Trans. Roy. Soc. London, vol. A247, pp. 529--551, April 1955.
\bibitem{b2} J. Clerk Maxwell, A Treatise on Electricity and Magnetism, 3rd ed., vol. 2. Oxford: Clarendon, 1892, pp.68--73.
\bibitem{b3} I. S. Jacobs and C. P. Bean, ``Fine particles, thin films and exchange anisotropy,'' in Magnetism, vol. III, G. T. Rado and H. Suhl, Eds. New York: Academic, 1963, pp. 271--350.
\bibitem{b4} K. Elissa, ``Title of paper if known,'' unpublished.
\bibitem{b5} R. Nicole, ``Title of paper with only first word capitalized,'' J. Name Stand. Abbrev., in press.
\bibitem{b6} Y. Yorozu, M. Hirano, K. Oka, and Y. Tagawa, ``Electron spectroscopy studies on magneto-optical media and plastic substrate interface,'' IEEE Transl. J. Magn. Japan, vol. 2, pp. 740--741, August 1987 [Digests 9th Annual Conf. Magnetics Japan, p. 301, 1982].
\bibitem{b7} M. Young, The Technical Writer's Handbook. Mill Valley, CA: University Science, 1989.
\end{thebibliography}

\end{document}